\documentclass[sigconf]{acmart}

\AtBeginDocument{%
  }

\usepackage{booktabs}
\usepackage{enumitem}
\usepackage[table]{xcolor}

\newcommand{\PathsTableStyle}{%
  \small
  \renewcommand{\arraystretch}{1.15}%
  \setlength{\tabcolsep}{4pt}%
}
\newcommand{\PathsDenseTableStyle}{%
  \footnotesize
  \renewcommand{\arraystretch}{1.15}%
  \setlength{\tabcolsep}{3pt}%
}
\copyrightyear{2026}
\acmYear{2026}
\setcopyright{cc}
\setcctype{by}
\acmConference[MM '26]{Proceedings of the 34th ACM International Conference on Multimedia}{November 10--14, 2026}{Rio de Janeiro, Brazil}
\acmBooktitle{Proceedings of the 34th ACM International Conference on Multimedia (MM '26), November 10--14, 2026, Rio de Janeiro, Brazil}
\acmDOI{10.1145/3767308.3835505}
\acmISBN{979-8-4007-2213-4/2026/11}

\begin{document}

\title[Paths for RGB-Event Video Person Re-Identification]{Paths: Prompt-aware Spatio-temporal Transformer with Hierarchical Multi-modal Fusion for RGB-Event Video Person Re-Identification}

\author{Yakun Huo}
\affiliation{%
  \institution{Dalian University of Technology}
  \city{Dalian}
  \country{China}}
\email{huoyakun@mail.dlut.edu.cn}

\author{Yingquan Wang}
\affiliation{%
  \institution{Dalian University of Technology}
  \city{Dalian}
  \country{China}}
\email{yingquan\_w95@mail.dlut.edu.cn}

\author{Yangyang Liu}
\affiliation{%
  \institution{Dalian University of Technology}
  \city{Dalian}
  \country{China}}
\email{yyliu@mail.dlut.edu.cn}

\author{Tianyu Yan}
\affiliation{%
  \institution{Dalian University of Technology}
  \city{Dalian}
  \country{China}}
\email{2981431354@mail.dlut.edu.cn}

\author{Yunzhi Zhuge}
\affiliation{%
  \institution{Dalian University of Technology}
  \city{Dalian}
  \country{China}}
\email{zgyz@dlut.edu.cn}

\author{Pingping Zhang}
\authornote{Corresponding author.}
\affiliation{%
  \institution{Dalian University of Technology}
  \city{Dalian}
  \country{China}}
\email{zhpp@dlut.edu.cn}

\author{Huchuan Lu}
\affiliation{%
  \institution{Dalian University of Technology}
  \city{Dalian}
  \country{China}}
\email{lhchuan@dlut.edu.cn}
\settopmatter{authorsperrow=4}

\renewcommand{\shortauthors}{Yakun Huo et al.}
\begin{abstract}
RGB-Event Video Person Re-Identification (RE-VReID) aims to retrieve specific person across non-overlapping cameras with complementary RGB videos and event streams.
However, existing methods often decouple spatial and temporal modeling, which limits their interaction.
In addition, global-level RGB-Event fusion fails to fully exploit fine-grained discriminative cues.
To address these issues, we propose Paths, a unified framework with spatio-temporal modeling and hierarchical multi-modal fusion for RE-VReID.
Specifically, we first design a Memory-Augmented Backbone (MAB) to maintain modality-specific identity prototypes for stable intra-modal representation learning.
Then, we propose a Prompt-aware Spatio-temporal Transformer (PST) to jointly model spatial and temporal cues within a unified Transformer.
Finally, we introduce a Hierarchical Multi-modal Fusion (HMF) to integrate RGB and event features at global and local levels.
With these modules, our framework can learn robust and discriminative representations for RE-VReID.
Extensive experiments on three public RE-VReID benchmarks including EvReID, MARS and iLIDS-VID, demonstrate the effectiveness of our proposed method.
The code is available at \url{https://github.com/Reflection0427/Paths}.
\end{abstract}



\maketitle
\begin{figure}[t]
  \centering
  \includegraphics[width=1.0\linewidth]{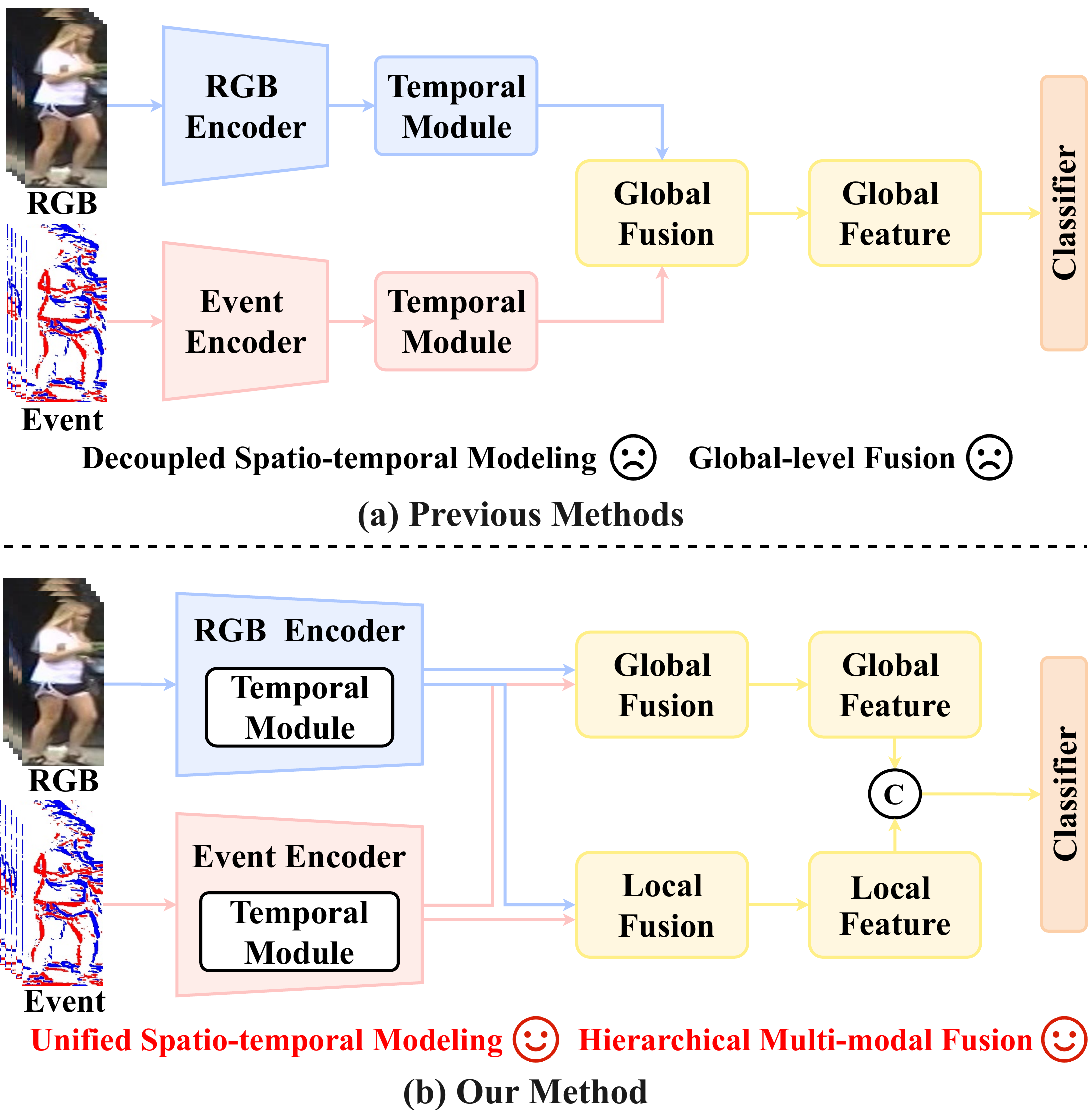}
  \caption{Comparison of different methods. (a) Previous methods decouple spatial and temporal modeling and rely on global-level fusion. (b) Our method unifies spatio-temporal modeling and employs hierarchical multi-modal fusion.}
  \label{fig:motivation}
\end{figure}
\section{Introduction}
Video-based Person Re-Identification (VReID) aims to retrieve specific person from video tracklets captured by non-overlapping cameras~\cite{li2018di,7780517,dai2019video,liu2021watching,liu2023long,liu2024video}.
It has important applications in intelligent surveillance and multimedia retrieval.
Most RGB-based VReID methods extract frame-level appearance features and aggregate them to capture temporal information~\cite{7780517,li2018di,eom2021video,wang2021pyramid,yang2026hihr}.
More recent methods further strengthen frame-to-frame interaction through graph reasoning or spatio-temporal attention~\cite{yan2020learning,liu2021,gu2020appearance,hou2021bicnet,liu2023deeply}.
However, RGB videos sampled at a fixed frame rate often contain substantial inter-frame redundancy~\cite{cao2023event,wang2026person}.
Event cameras provide a complementary modality by asynchronously recording pixel-level brightness changes.
Such asynchronous sensing is particularly sensitive to motion variations and temporal changes~\cite{gallego2022event}.
This complementarity enables RGB-Event Video Person Re-Identification to jointly exploit appearance cues from RGB videos and
motion cues from event streams.
Existing studies have explored sparse-dense complementary learning~\cite{cao2023event}, cross-modality and temporal collaboration~\cite{li2025eventbasedcmtc}, and attribute-guided semantic interaction~\cite{wang2026person}.
These studies fully demonstrate the benefit of jointly exploiting RGB and event information for video person representation learning.
\begin{figure*}[!t]
  \centering
  \includegraphics[width=\textwidth]{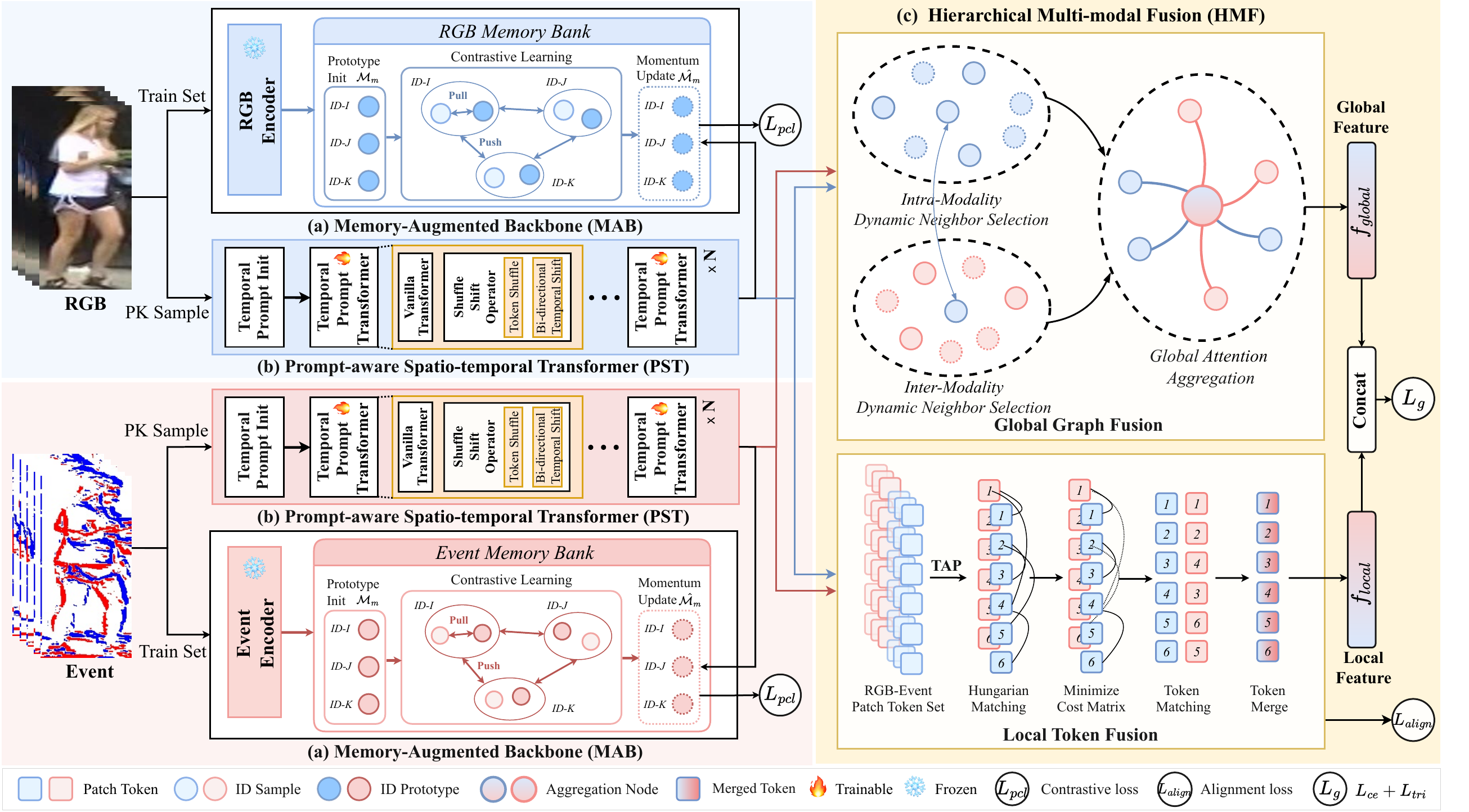}
  \caption{Overview of the proposed framework, including Memory-Augmented Backbone (MAB), Prompt-aware Spatio-temporal Transformer (PST), and Hierarchical Multi-modal Fusion (HMF).}
  \label{fig:framework}
\end{figure*}

Despite these advances, previous RE-VReID methods still face two key limitations.
As shown in Fig.~\ref{fig:motivation} (a), previous methods usually extract RGB and event frame features separately and fuse them only at the global or video level via temporal average pooling~\cite{cao2023event}.
Although this strategy can combine complementary information from the two modalities, it overlooks the modality differences and the fine-grained spatial misalignment between RGB and event streams.
Moreover, we further observe a common bottleneck in the spatio-temporal modeling of existing methods.
Spatial feature extraction and temporal modeling are decoupled into two sequential stages.
Such a two-stage pipeline restricts the interaction between spatial semantics and temporal dynamics, leading to insufficient spatio-temporal representations.

Based on the above analysis, we rethink the RE-VReID paradigm from two perspectives: (i) spatial and temporal modeling should be unified rather than sequentially decoupled;
(ii) cross-modal fusion should interact at both global and local levels to bridge the modality gap. %
Motivated by these insights, we propose Paths, a novel framework with spatio-temporal modeling and hierarchical cross-modal fusion for RE-VReID.
As illustrated in Fig.~\ref{fig:motivation} (b), our framework consists of three components: Memory-Augmented Backbone (MAB), Prompt-aware Spatio-temporal Transformer (PST), and Hierarchical Multi-modal Fusion (HMF).
Specifically, MAB maintains modality-specific identity prototypes for stable intra-modal representation learning.
PST leverages learnable temporal prompts to jointly capture spatial semantics and temporal dynamics within a unified Transformer.
Furthermore, to bridge the modality gap, HMF integrates RGB and event features at both global and local levels.
By integrating these components, our framework can learn more robust and discriminative identity representations for RE-VReID.
Extensive experiments on three public RE-VReID benchmarks demonstrate the effectiveness of our method.

In summary, our main contributions are as follows:
\begin{itemize}[topsep=0pt, partopsep=0pt, itemsep=0pt, parsep=0pt]
  \item We propose Paths, a novel framework with spatio-temporal modeling and hierarchical multi-modal fusion for RE-VReID.
  \item We design a Memory-Augmented Backbone (MAB) to maintain modality-specific identity prototypes for stable intra-modal representation learning.
  \item We develop a Prompt-aware Spatio-temporal Transformer (PST) to jointly model spatial and temporal cues.
  \item We introduce a Hierarchical Multi-modal Fusion (HMF) to fuse RGB and event features at both global and local levels.
  \item Extensive experiments on three benchmark datasets demonstrate the effectiveness of our proposed framework.
\end{itemize}

\section{Related Work}
\subsection{RGB-Event Video Person ReID}
RGB-Event Video Person Re-Identification (RE-VReID) aims to retrieve specific person by jointly exploiting appearance cues from RGB videos and motion information from event streams.
Existing studies mainly focus on event representation enhancement, RGB-Event complementary learning, and semantic-guided cross-modal interaction
~\cite{cao2023event,chen2025hierarchical,li2025eventbasedcmtc,Tan2025SpectrumguidedFE,wang2026person}.
For event representation enhancement, SFE-Net~\cite{Tan2025SpectrumguidedFE} combines frequency-domain filtering and multi-granularity spatial enhancement to suppress event noise.
HSCPS-Net~\cite{chen2025hierarchical} uses global and local proxy memories to model multi-level semantics and an event inversion model to capture fine-grained event information.
For RGB-Event complementary learning, SDCL~\cite{cao2023event} uses separate CNN and SNN branches to encode RGB and event features, and further performs cross-modal alignment.
CMTC~\cite{li2025eventbasedcmtc} generates auxiliary representations from raw events and enhances their complementarity through modality and temporal collaboration.
More recently, TriPro-ReID~\cite{wang2026person} uses person attributes as intermediate semantic information and introduces attribute and cross-modal prompts to facilitate collaborative learning between RGB and event.
These methods improve RGB-Event representation learning from different perspectives.
Despite continuous progress, existing fusion methods still rely on global-level representations.
In contrast, our method jointly models spatial and temporal information within a unified Transformer and performs hierarchical RGB-Event fusion at both global and local levels.
\subsection{Prompt Learning}
Prompt learning adapts visual models by introducing a small number of learnable prompt tokens
~\cite{zhou2022learningprompt,jia2022visualprompt,khattak2023maple}.
In person ReID, CLIP-ReID~\cite{li2023clip} learns identity-specific textual tokens to encode identity semantics and transfers the resulting text knowledge to visual representation learning.
LATex~\cite{hu2025latex} exploits attribute-based textual knowledge through structured textual prompts.
APC~\cite{wang2026unique} constructs identity-relevant adaptive prompt composition to improve the discriminative and generalizable person ReID.
$\pi$-VL~\cite{lin2025exploringpartinformedvisuallanguagelearning} further incorporates part-aware prompts, allowing local body regions to interact with semantic descriptions for fine-grained identity learning.
For video-based person ReID, TF-CLIP~\cite{yu2024tf} introduces a video-level prompt to dynamically update the identity-specific memory for robust representation learning.
For RE-VReID, TriPro-ReID~\cite{wang2026person} introduces attribute and cross-modal prompts to inject person semantics and facilitate interaction between RGB and event representations.
Beyond the above works, prompt learning has also been extended to temporal modeling.
For example, VoP~\cite{huang2023vop} introduces learnable video prompts to incorporate temporal information across frames, while STOP~\cite{liu2025stop} generates dynamic spatio-temporal prompts according to video variations.
These studies demonstrate the potential of prompts for semantic guidance and video representation learning.
However, they mainly use prompts for representation adaptation or temporal aggregation, without explicitly using them to propagate temporal information across frames.
In contrast, our method introduces learnable temporal prompts for spatio-temporal modeling. It allows spatial and temporal cues to be jointly modeled within a unified Transformer.
\subsection{Multi-modal Fusion}
Multi-modal fusion leverages complementary information from heterogeneous modalities to improve feature robustness~\cite{li2026cadtrack,li2026ragtrack,wang2026sdreid,hu2025latex,yang2026sasvpreid,nguyen2025agvpreid,hambarde2026vreidxfd}.
Recent studies increasingly focus on fine-grained interactions among multi-modal representations.
For example, TOP-ReID~\cite{wang2024top} introduces token permutation to exchange local information across different spectra, with complementary reconstruction for cross-modal learning.
IEEE~\cite{wang2022interact} exchanges information across modalities and incorporates fine-grained local cues into multi-modal representations.
Magic Tokens~\cite{zhang2024magictokens} selects object-centric tokens and performs hierarchical masked aggregation within and across modalities.
AIO~\cite{li2024all} projects heterogeneous modalities into a unified representation space for multi-modal ReID.
DeMo~\cite{wang2025demo} decouples multi-modal features and adaptively balances them through a mixture-of-experts architecture.
DMPT~\cite{lin2025dmpt} introduces modality-aware prompts to exchange complementary information across modalities.
Uni-Prompt ReID~\cite{ha2025multimodal} utilizes modality and platform-aware prompts for representation learning across heterogeneous inputs.
MambaPro~\cite{wang2025mambapro} combines synergistic prompts with Mamba aggregation for cross-modal interaction.
IDEA~\cite{wang2025idea} incorporates text semantics into multi-modal representation learning and uses deformable aggregation to capture complementary global and local information.
More recently, Signal~\cite{liu2026signal} selects informative tokens for cross-modal interaction and aligns multi-modal representations at both global and local levels.
These methods demonstrate the effectiveness of fine-grained interaction and hierarchical aggregation for multi-modal representation learning.
Different from these methods, our method performs hierarchical RGB-Event fusion at both global and local levels.
\section{Methodology}
As shown in Fig.~\ref{fig:framework}, our proposed framework comprises three main components: Memory-Augmented Backbone (MAB), Prompt-aware Spatio-temporal Transformer (PST), and Hierarchical Multi-modal Fusion (HMF).
Details of the each component are as follows.
\subsection{Memory-Augmented Backbone}
Existing RE-VReID methods~\cite{cao2026learning} mainly rely on
samples within the current mini-batch for representation learning,
making the learned features sensitive to the batch sampling.
Inspired by X-ReID~\cite{yu2026xreid}, we propose a
Memory-Augmented Backbone (MAB) to maintain modality-specific
identity prototypes across mini-batches.

\noindent\textbf{Multi-modal Feature Encoding.}
In this work, the training set is denoted as $\mathcal{D}=\{(V_i^{r},V_i^{e},y_i)\}_{i=1}^{I}$.
Here, $V_i^{r}$ and $V_i^{e}$ are the RGB and event video sequences of the $i$-th training sample, respectively.
$y_i\in\{1,\ldots,Y\}$ is the corresponding identity label.
$I$ is the number of training samples.
$Y$ is the total number of training identities.
$m\in\{r,e\}$ is the modality, where $r$ and $e$ represent RGB and event, respectively.
For each modality, we uniformly sample $T$ frames from the video sequence, denoted as $V^{m}=\{I_t^{m}\}_{t=1}^{T}$, where $I_t^{m}$ is the $t$-th frame image.
Each frame $I_t^{m}$ is divided into $N$ patches and fed into the visual encoder $\Phi^{m}(\cdot)$:
\setlength{\abovedisplayskip}{3pt}
\setlength{\belowdisplayskip}{3pt}
\begin{equation}
\left[
\hat{f}_t^{m};
f_{t,1}^{m};
\ldots;
f_{t,N}^{m}
\right]
=
\Phi^{m}(I_t^{m}),
\end{equation}
where $\hat{f}_t^{m}\in\mathbb{R}^{D}$ is the class token, $f_{t,n}^{m}\in\mathbb{R}^{D}$ is the $n$-th patch token, $n\in\{1,\ldots,N\}$, and $D$ is the feature dimension.
We then apply a Temporal Average Pooling (TAP) to the class tokens of all frames to obtain the sequence-level representation:
\begin{equation}
\bar{f}^{m}
=
\frac{1}{T}
\sum_{t=1}^{T}
\hat{f}_t^{m}.
\end{equation}
The resulting $\bar{f}^{m}$ provides a compact representation of the entire video sequence and is used for identity-level prototype learning.

\noindent\textbf{Prototype Memory Learning.}
To provide stable cross-batch identity references, we construct a modality-specific prototype memory for each modality, as follows:
\begin{equation}
\mathcal{M}^{m}
=
\left\{
p_y^{m}
\right\}_{y=1}^{Y},
\end{equation}
where $p_y^m\in\mathbb{R}^{D}$ is the prototype of identity $y$ under modality $m$, and $\mathcal{M}^{m}\in\mathbb{R}^{Y\times D}$ stores each training identity prototype.
Before training, we initialize the prototype memory using the frozen pre-trained visual encoder.
Specifically, we extract and aggregate the sequence-level representations of all training samples.
For identity $y$ under modality $m$, its prototype is initialized as follows:
\begin{equation}
p_y^{m}
=
\frac{1}{|\mathcal{D}_y^m|}
\sum_{q\in\mathcal{D}_y^m}
\bar{f}_q^m,
\end{equation}
where $\mathcal{D}_y^m$ is the set of training samples belonging to identity $y$ under modality $m$, and $\bar{f}_q^m$ is the sequence-level representation of the sample $q$.
After initialization, the prototypes are progressively updated with the current mini-batch features using a momentum update strategy, as follows:
\begin{equation}
p_y^{m}
\leftarrow
\mu p_y^{m}
+
(1-\mu)\tilde{f}_y^{m},
\end{equation}
where $\mu\in[0,1]$ is the momentum coefficient.
For identity $y$, the mean representation of its samples in the current mini-batch is computed as:
\begin{equation}
\tilde{f}_y^{m}
=
\frac{1}{|\mathcal{B}_y^{m}|}
\sum_{q\in\mathcal{B}_y^{m}}
\bar{f}_q^{m},
\end{equation}
where $\mathcal{B}_y^{m}$ is the set of samples belonging to identity $y$ under modality $m$ in the current mini-batch.

\noindent\textbf{Prototypical Contrastive Learning.}
To guide intra-modal representation learning, we further introduce a prototypical contrastive learning loss.
For a sequence representation $\bar{f}^{m}$ with identity label $y$, the loss is defined as:
\begin{equation}
\mathcal{L}_{pcl}^{m}
=
-\log
\frac{
\exp\left(
\operatorname{sim}
\left(
\bar{f}^{m},
p_y^{m}
\right)/\tau
\right)
}{
\sum_{y'=1}^{Y}
\exp\left(
\operatorname{sim}
\left(
\bar{f}^{m},
p_{y'}^{m}
\right)/\tau
\right)
},
\end{equation}
where $\tau$ is the temperature parameter and $\operatorname{sim}(\cdot,\cdot)$ is the cosine similarity.
Here, $p_y^{m}$ is the prototype corresponding to the ground-truth identity, while $p_{y'}^{m}$ enumerates all identity prototypes in modality $m$.
Therefore, each sequence representation is compared against the prototypes of all training identities rather than only the current mini-batch.
By pulling each sequence representation toward its corresponding prototype while separating it from other identity prototypes, MAB reduces intra-identity variation and improves the discrimination of intra-modal representations.
\begin{figure}[t]
  \centering
  \includegraphics[width=\linewidth]{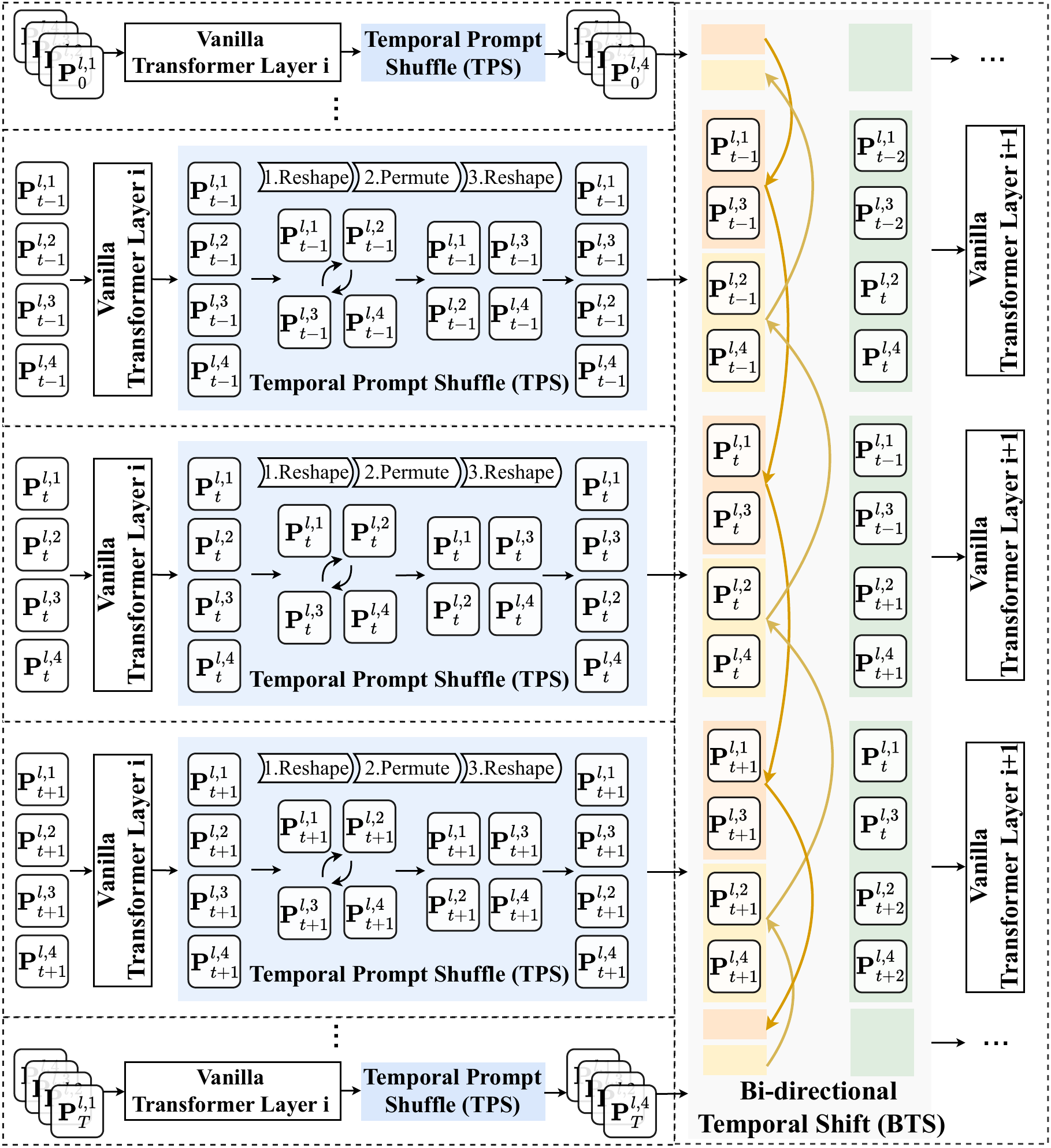}
  \caption{Detailed structure of the proposed Prompt-aware Spatio-temporal Transformer (PST).}
  \label{fig:PST}
\end{figure}
\subsection{Prompt-aware Spatio-temporal Transformer}
Existing RE-VReID methods commonly separate spatial representation learning from temporal modeling, limiting the interaction between spatial semantics and temporal information.
To alleviate this issue, we propose a Prompt-aware Spatio-temporal Transformer (PST), which leverages learnable temporal prompts to jointly capture spatial semantics and temporal dynamics.
As shown in Fig.~\ref{fig:PST}, it consists of Temporal Prompt Shuffle (TPS) and Bidirectional Temporal Shift (BTS).
The TPS reorganizes the temporal prompt groups, while the BTS exchanges them between adjacent frames before they are fed into the next Transformer layer.

\noindent\textbf{Temporal Prompt Initialization.}
For each Transformer layer, we introduce $G$ learnable temporal prompt tokens for every frame.
At the $l$-th layer, the temporal prompts are denoted as $P^{l}\in\mathbb{R}^{T\times G\times D}$, where $T$ is the number of frames, $G$ is the number of prompt tokens, and $D$ is the feature dimension.
For the $t$-th frame, the prompt tokens are represented as $\{P_{t}^{l,g}\}_{g=1}^{G}$.
$P_t^{l,g}$ is the $g$-th temporal prompt group of the $t$-th frame at the $l$-th Transformer layer.
Following the notation in the MAB, the input tokens of the $t$-th frame at layer $l$ are constructed as:
\begin{equation}
F_t^{m,l}
=
\left[
\hat{f}_t^{m,l};
f_{t,1}^{m,l};
\ldots;
f_{t,N}^{m,l};
P_{t}^{l};
\ldots;
P_{t}^{l,G}
\right]
\in
\mathbb{R}^{(1+N+G)\times D},
\end{equation}
where $\hat{f}_t^{m,l}\in\mathbb{R}^{D}$ and $f_{t,n}^{m,l}\in\mathbb{R}^{D}$ denote the class token and the $n$-th patch token of modality $m$ at the $l$-th layer, respectively.
The temporal prompts interact with the class and patch tokens through Transformer layers, allowing temporal information to participate directly in spatial representation learning.
Although temporal prompts interact with spatial tokens within each frame, direct information exchange across different frames is still limited.
We therefore introduce the temporal prompt shuffle and bidirectional temporal shift to propagate the prompt information along the temporal dimension.

\noindent\textbf{Temporal Prompt Shuffle.}
To reorganize the prompts before temporal propagation, we divide the $G$ prompts into $g$ groups, each containing $n_g=G/g$ tokens.
The prompt tensor is first reshaped as:
\begin{equation}
P_{\mathit{res}}^{l}
=
\operatorname{Reshape}(P^{l})
\in
\mathbb{R}^{T\times g\times n_g\times D}.
\end{equation}
We then exchange the group and token dimensions:
\begin{equation}
P_{\mathit{perm}}^{l}
=
\operatorname{Permute}
\left(
P_{\mathit{res}}^{l}
\right)
\in
\mathbb{R}^{T\times n_g\times g\times D}.
\end{equation}
Finally, the permuted tensor is reshaped back to the original layout:
\begin{equation}
\hat{P}^{l}
=
\operatorname{Reshape}
\left(
P_{\mathit{perm}}^{l}
\right)
\in
\mathbb{R}^{T\times G\times D}.
\end{equation}
The above operation redistributes prompts across different groups, so that different temporal spans can participate in the subsequent temporal exchange.

\noindent\textbf{Bidirectional Temporal Shift.}
After the temporal prompt shuffle, we split the prompts into two equal subsets along the prompt dimension: the forward group $\hat{P}_{fwd}$ and the backward group $\hat{P}_{bwd}$, each containing half prompts:
\begin{equation}
\hat{P}^{l}
=
\left[
\hat{P}_{\mathit{fwd}}^{l};
\hat{P}_{\mathit{bwd}}^{l}
\right],
\end{equation}
where
$\hat{P}_{\mathit{fwd}}^{l},
\hat{P}_{\mathit{bwd}}^{l}
\in\mathbb{R}^{T\times \frac{G}{2}\times D}$.
The forward subset is shifted from the previous frame to the current frame:
\begin{equation}
\tilde{P}_{\mathit{fwd},t}^{l}
=
\begin{cases}
\hat{P}_{\mathit{fwd},t}^{l}, & t=1,\\
\hat{P}_{\mathit{fwd},t-1}^{l}, & 1<t\leq T,
\end{cases}
\end{equation}
while the backward subset is shifted from the next frame:
\begin{equation}
\tilde{P}_{\mathit{bwd},t}^{l}
=
\begin{cases}
\hat{P}_{\mathit{bwd},t+1}^{l}, & 1\leq t<T,\\
\hat{P}_{\mathit{bwd},t}^{l}, & t=T.
\end{cases}
\end{equation}
The two shifted subsets are then concatenated:
\begin{equation}
\tilde{P}^{l}
=
\left[
\tilde{P}_{\mathit{fwd}}^{l};
\tilde{P}_{\mathit{bwd}}^{l}
\right]
\in
\mathbb{R}^{T\times G\times D}.
\end{equation}
In this way, the prompts associated with each frame receive information from its preceding and succeeding frames.
By combining prompt-based spatial interaction with bidirectional cross-frame propagation, the proposed PST allows spatial and temporal cues to be jointly modeled within the same Transformer backbone without introducing an additional temporal encoder.
\subsection{Hierarchical Multi-modal Fusion}
Existing RE-VReID methods mainly focus on global-level feature fusion, while fine-grained interactions between RGB and event remain insufficiently explored.
To address this limitation, we propose a Hierarchical Multi-modal Fusion (HMF), which integrates RGB and event features at both global and local levels.
It consists of Global Graph Fusion (GGF) and Local Token Fusion (LTF).
\begin{figure}[t]
\centering
\includegraphics[width=\linewidth]{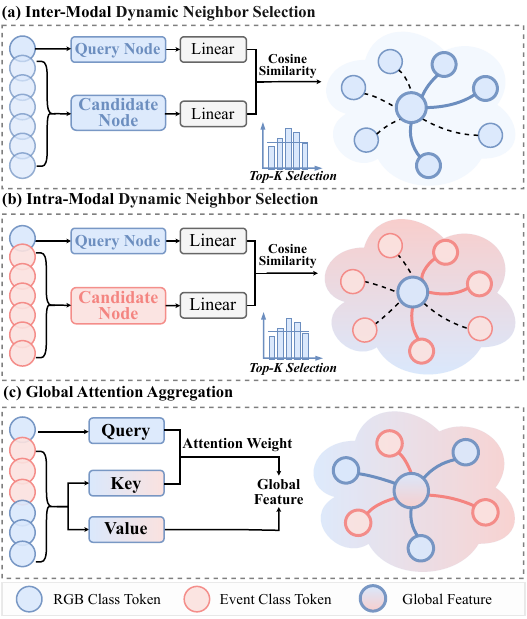}
\vspace{-15pt}
\caption{Illustration of our Global Graph Fusion.}
\label{fig:Global}
\end{figure}

\noindent\textbf{Global Graph Fusion (GGF).}
As shown in Fig.~\ref{fig:Global}, GGF models cross-modal and cross-temporal relationships among frame-level representations.
We regard the RGB and event frame features as nodes of a dynamic graph.
Let $u_t^{m}\in\mathbb{R}^{D}$ denote the frame-level representation of the $t$-th frame under modality $m$, obtained from the class token $\hat{f}_t^{m,L}$ of the last PST layer.
For each query node $u_t^{m}$, GGF first selects informative intra-modal and inter-modal neighbors and then aggregates them through a multi-head attention.

\noindent\textit{(1) Dynamic Neighbor Selection.}
For a query node $u_t^{m}$, we construct an intra-modal candidate set from the remaining frames of the same modality, as follows:
\begin{equation}
\mathcal{C}_{t}^{\mathit{intra}}
=
\left\{
u_s^{m}
\mid
s\neq t,\;
1\leq s\leq T
\right\},
\end{equation}
and an inter-modal candidate set from the other modality as:
\begin{equation}
\mathcal{C}_{t}^{\mathit{inter}}
=
\left\{
u_s^{\bar{m}}
\mid
1\leq s\leq T
\right\},
\end{equation}
where $\bar{m}$ denotes the modality complementary to $m$, i.e., $\bar{r}=e$ and $\bar{e}=r$.
This construction captures both intra-modal temporal context and
cross-modal complementary information.
The relevance of candidate node $c_i$ to query node $u_t^{m}$ is
computed by:
\begin{equation}
a_{ti}
=
\frac{
\left(W_q u_t^{m}\right)^{\top}
\left(W_k c_i\right)
}{
\gamma\sqrt{D/N_h}
},
\end{equation}
where $W_q$ and $W_k$ are learnable projection matrices.
$N_h$ is the number of attention heads, and $\gamma$ is a temperature parameter.
We independently select the Top-$K$ candidates from the intra-modal and inter-modal candidate sets:
\begin{equation}
\mathcal{N}_{t}^{\mathit{intra}}
=
\operatorname{TopK}
\left(
\mathcal{C}_{t}^{\mathit{intra}},
a_{ti},
K
\right),
\end{equation}
\begin{equation}
\mathcal{N}_{t}^{\mathit{inter}}
=
\operatorname{TopK}
\left(
\mathcal{C}_{t}^{\mathit{inter}},
a_{ti},
K
\right).
\end{equation}
The local graph centered at $u_t^{m}$ is then constructed as:
\begin{equation}
\mathcal{G}_{t}^{m}
=
\left\{
u_t^{m}
\right\}
\cup
\mathcal{N}_{t}^{\mathit{intra}}
\cup
\mathcal{N}_{t}^{\mathit{inter}}.
\end{equation}
The dynamic neighbor selection filters out less relevant frames and retains the most informative temporal and cross-modal cues for each query node, improving the effective of feature aggregation.

\noindent\textit{(2) Global Attention Aggregation.}
Given the constructed graph $\mathcal{G}_{t}^{m}$, we aggregate information from the selected neighbors using a multi-head attention.
For the $h$-th attention head, the query, key, and value features are computed as:
\begin{equation}
q_t^{(h)}
=
W_q^{(h)}u_t^{m},
\qquad
k_j^{(h)}
=
W_k^{(h)}g_j,
\qquad
v_j^{(h)}
=
W_v^{(h)}g_j,
\end{equation}
where $g_j\in\mathcal{G}_{t}^{m}$ denotes a node in the constructed graph.
The corresponding attention weight is defined as:
\begin{equation}
\omega_{tj}^{(h)}
=
\frac{
\exp\left(
\frac{
(q_t^{(h)})^{\top}k_j^{(h)}
}{
\sqrt{D/N_h}
}
\right)
}{
\sum\limits_{g_s\in\mathcal{G}_{t}^{m}}
\exp\left(
\frac{
(q_t^{(h)})^{\top}k_s^{(h)}
}{
\sqrt{D/N_h}
}
\right)
}.
\end{equation}
The aggregated representation of the $h$-th head is:
\begin{equation}
z_t^{(h)}
=
\sum_{g_j\in\mathcal{G}_{t}^{m}}
\omega_{tj}^{(h)}v_j^{(h)}.
\end{equation}
The outputs of all attention heads are concatenated and projected to obtain the enhanced node representation:
\begin{equation}
\hat{u}_t^{m}
=
W_{\mathit{out}}
\left[
z_t^{(1)};
\ldots;
z_t^{(N_h)}
\right].
\end{equation}
Finally, the enhanced frame-level representations from different modalities are aggregated into a global representation:
\begin{equation}
f_{\mathit{global}}
=
W_{\mathit{agg}}
\left(
\frac{1}{2T}
\sum_{m\in\{r,e\}}
\sum_{t=1}^{T}
\hat{u}_t^{m}
\right),
\end{equation}
where $W_{\mathit{agg}}$ is a learnable projection matrix.
By dynamically selecting and aggregating informative intra-modal and inter-modal neighbors, GGF models cross-temporal dependencies and complementary RGB-Event cues,
producing a unified global representation.

\noindent\textbf{Local Token Fusion (LTF).}
Although GGF captures cross-modal and cross-temporal features at the global level, such global fusion cannot resolve the fine-grained spatial misalignment between RGB and event streams.
To establish local correspondences and integrate complementary token-level information, we introduce a Local Token Fusion (LTF).
Specifically, LTF consists of two stages: \emph{Optimal Token Matching} and \emph{Adaptive Gated Fusion}.

\noindent\textit{(1) Optimal Token Matching.}
We use the patch tokens from the last PST layer for local cross-modal matching.
For each patch position $n$, a Temporal Average Pooling (TAP) is first applied across all frames:
\begin{equation}
\bar{f}_n^{m}
=
\frac{1}{T}
\sum_{t=1}^{T}
f_{t,n}^{m,L},
\end{equation}
where $f_{t,n}^{m,L}\in\mathbb{R}^{D}$ is the $n$-th patch token of the $t$-th frame from the last PST layer.
$\bar{f}_n^{m}\in\mathbb{R}^{D}$ is the sequence-level representation of the $n$-th local token under
modality $m$.
We then compute the cosine similarity between the RGB and event tokens:
\begin{equation}
S_{ij}
=
\frac{
\left(\bar{f}_i^{r}\right)^{\top}
\bar{f}_j^{e}
}{
\left\|\bar{f}_i^{r}\right\|_2
\left\|\bar{f}_j^{e}\right\|_2
},
\end{equation}
where $i,j\in\{1,\ldots,N\}$.
The matching cost is defined as:
\begin{equation}
C_{ij}
=
1-S_{ij}.
\end{equation}
To establish explicit one-to-one correspondences between RGB and event tokens, we employ the Hungarian algorithm~\cite{Hungarian} to obtain the minimum-cost assignment:
\begin{equation}
\pi^{*}
=
\arg\min_{\pi}
\sum_{i=1}^{N}
C_{i,\pi(i)},
\end{equation}
where $\pi(i)$ denotes the event token assigned to the $i$-th RGB token.
Based on the optimal assignment, the matched token pair is denoted as
$\left(\bar{f}_i^{r},
\bar{f}_{\pi^{*}(i)}^{e}\right)$.
We define the cross-modal alignment loss as:
\begin{equation}
\mathcal{L}_{\mathit{align}}
=
\frac{1}{N}
\sum_{i=1}^{N}
\left[
1-
\operatorname{cos}
\left(
\bar{f}_i^{r},
\bar{f}_{\pi^{*}(i)}^{e}
\right)
\right].
\end{equation}

\noindent\textit{(2) Adaptive Gated Fusion.}
Although the optimal matching establishes cross-modal correspondences, RGB and event tokens contribute differently to the fused representation.
Therefore, we introduce an adaptive fusion gate to dynamically balance the information from each matched token pair.
For the $i$-th matched pair, the fusion coefficient is computed as:
\begin{equation}
g_i
=
\sigma
\left(
W_2
\delta
\left(
W_1
\left[
\bar{f}_i^{r};
\bar{f}_{\pi^{*}(i)}^{e}
\right]
\right)
\right),
\end{equation}
where $W_1$ and $W_2$ are learnable projection matrices.
$\delta(\cdot)$ is the ReLU activation function, and $\sigma(\cdot)$ is the Sigmoid function.
The adaptively fused local token is formulated as:
\begin{equation}
\tilde{f}_i
=
\operatorname{LN}
\left(
g_i\bar{f}_i^{r}
+
(1-g_i)
\bar{f}_{\pi^{*}(i)}^{e}
\right),
\end{equation}
where $\operatorname{LN}(\cdot)$ is layer normalization.
Finally, we use a learnable pooling query $q_{\mathit{pool}}\in\mathbb{R}^{D}$ to aggregate the fused local
tokens, where $\beta_i$ is the normalized attention weight of the $i$-th fused token computed with $q_{\mathit{pool}}$.
The final local representation is obtained as:
\begin{equation}
f_{\mathit{local}}
=
\sum_{i=1}^{N}
\beta_i\tilde{f}_i.
\end{equation}
Through the above operations, LTF can establish fine-grained correspondences between local patch tokens from the two modalities and then fuses the matched tokens.

\subsection{Objective Functions}
As shown in Fig.~\ref{fig:framework}, our proposed framework is jointly optimized with four objectives: the cross-entropy loss $\mathcal{L}_{\mathit{ce}}$~\cite{szegedy2016rethinking}, the triplet loss $\mathcal{L}_{\mathit{tri}}$~\cite{hermans2017defense},, the prototypical contrastive loss $\mathcal{L}_{\mathit{pcl}}$, and the cross-modal alignment loss $\mathcal{L}_{\mathit{align}}$.
%
%
The prototypical contrastive loss is computed over two modalities as:
\begin{equation}
\mathcal{L}_{pcl}
=
\mathcal{L}_{pcl}^{r}
+
\mathcal{L}_{pcl}^{e},
\end{equation}
and the overall training objective is formulated as:
\begin{equation}
\mathcal{L}_{\mathit{total}}
=
\mathcal{L}_{\mathit{ce}}
+
\mathcal{L}_{\mathit{tri}}
+
\lambda_{1}\mathcal{L}_{\mathit{pcl}}
+
\lambda_{2}\mathcal{L}_{\mathit{align}},
\end{equation}
where $\lambda_{1}$ and $\lambda_{2}$ are the weighting coefficients of the prototypical contrastive loss and cross-modal alignment loss, respectively.
\begin{table}[!t]
  \caption{Comparison with different methods on EvReID.}
  \label{tab:sota_evreid}
  \centering
  \PathsTableStyle
  \begin{tabular}{cc*{4}{c}}
    \noalign{\hrule height 1pt}
    Method & Backbone & mAP & R-1 & R-5 & R-10 \\
    \hline
    OSNet~\cite{zhou2019omni} & ResNet50 & 23.7 & 49.1 & 65.4 & 72.3 \\
    TCLNet~\cite{hou2020temporal} & ResNet50 & 55.8 & 77.4 & 89.0 & 93.1 \\
    AP3D~\cite{gu2020appearance} & ResNet50 & 66.9 & 86.5 & \underline{95.6} & \underline{96.5} \\
    MGH~\cite{yan2020learning} & ResNet50 & 43.2 & 70.9 & 89.4 & 92.7 \\
    STMN~\cite{eom2021video} & ResNet50 & 42.1 & 73.8 & -- & -- \\
    PSTA~\cite{wang2021pyramid} & ResNet50 & 68.2 & 82.3 & 90.8 & 94.5 \\
    GRL~\cite{liu2021watching} & ResNet50 & 38.9 & 62.6 & 78.0 & 83.6 \\
    BiCnet-TKS~\cite{hou2021bicnet} & ResNet50 & 50.8 & 80.5 & 89.62 & 92.45 \\
    SINet~\cite{bai2022salient} & ResNet50 & 50.2 & 77.4 & 92.8 & 96.2 \\
    SDCL~\cite{cao2023event} & ResNet50 & 54.2 & 69.3 & 83.8 & 87.1 \\
    DCCT~\cite{liu2023deeply} & ViT-B/16 & 24.6 & 42.7 & 64.9 & 75.5 \\
    CLIP-ReID~\cite{li2023clip} & CLIP-B/16 & 49.2 & 73.0 & 85.5 & 91.5 \\
    TF-CLIP~\cite{yu2024tf} & CLIP-B/16 & 56.9 & 78.6 & 91.8 & 94.3 \\
    DeMo~\cite{wang2025demo} & CLIP-B/16 & 59.4 & 75.7 & 90.1 & 92.8 \\
    CLIMB-ReID~\cite{yu2025climb} & CLIP-B/16 & 68.3 & 85.2 & 92.8 & 95.8 \\
    TriPro-ReID~\cite{wang2026person} & CLIP-B/16 & 69.3 & 88.6 & 94.3 & 95.4 \\
    \hline
    \textbf{Paths}$^{\ast}$ & CLIP-B/16 & \underline{71.1} & \underline{89.3} & 94.6 & \underline{96.5} \\
    \rowcolor[gray]{0.92}
    \textbf{Paths}$^{\dagger}$ & DINOv3 & \textbf{73.6} & \textbf{90.8} & \textbf{96.2} & \textbf{97.8} \\
    \noalign{\hrule height 1pt}
  \end{tabular}
\end{table}
\section{Experiments}
\subsection{Datasets and Evaluation Metrics}
To evaluate the effectiveness of our proposed method, we conduct experiments on three RE-VReID benchmarks: the large-scale real-world EvReID dataset~\cite{wang2026person} and two simulated datasets, MARS~\cite{zheng2016mars} and iLIDS-VID~\cite{wang2014person}.
Following previous works, we adopt the mean Average Precision (mAP) and Cumulative Matching haracteristics (CMC) at Rank-K (K= 1,5,10) as our evaluation metrics.
\subsection{Implementation Details}
Our model is implemented in PyTorch and trained on a single NVIDIA A100 GPU with 80\,GB of memory.
We evaluate two pre-trained visual backbones, CLIP-B/16~\cite{dosovitskiy2020image} and DINOv3~\cite{simeoni2025dinov3}.
For each backbone, the RGB and event encoders are initialized with the corresponding pre-trained weights.
For each tracklet, we uniformly sample $T=8$ frames and resize each frame to $256\times128$.
Each mini-batch contains 16 identities and 4 paired RGB-Event tracklets per identity, resulting in 64 paired tracklets.
For data augmentation, we apply random horizontal flipping, padding followed by random cropping, and random erasing~\cite{zhong2017randomerasingdataaugmentation}.
The model is optimized using Adam~\cite{Kingma2014AdamAM} with a base learning rate of $5\times10^{-6}$.
We apply linear learning-rate warm-up for the first 10 epochs,followed by cosine annealing.
The model is trained for 80 epochs in total.
For MAB, the momentum coefficient is set to $\mu = 0.2$.
For PST, the number of temporal prompt tokens is set to $G = 8$, and the number of prompt groups is set to $g = 2$.
For GGF, the numbers of selected intra-modal and inter-modal neighbors
are both set to 3, i.e.,
$K_{\mathit{intra}}=K_{\mathit{inter}}=3$.
The loss weights are set to $\lambda_{1}=0.3$ and $\lambda_{2}=0.1$ for the prototypical contrastive loss and cross-modal alignment loss, respectively.
\subsection{Comparison with State-of-the-art Methods}
To evaluate the effectiveness of the proposed framework, we conduct a comprehensive comparison with SOTA methods.

\noindent\textbf{Results on EvReID.}
As shown in Tab.~\ref{tab:sota_evreid}, our method achieves competitive performance across all evaluation metrics on EvReID.
With CLIP-B/16, \textbf{Paths}$^{\ast}$ obtains 71.1\% mAP and 89.3\% Rank-1.
Compared with TriPro-ReID~\cite{wang2026person}, \textbf{Paths}$^{\ast}$ improves mAP and Rank-1 by 1.8\% and 0.7\%, respectively.
When using DINOv3, \textbf{Paths}$^{\dagger}$ further improves mAP to 73.6\% and Rank-1 to 90.8\%.
These results demonstrate the strong overall performance of our method.

\noindent\textbf{Results on MARS.}
As shown in Tab.~\ref{tab:sota_mars}, \textbf{Paths}$^{\ast}$ achieves 88.1\% mAP and 91.3\% Rank-1 with CLIP-B/16.
With DINOv3, \textbf{Paths}$^{\dagger}$ reaches 89.4\% mAP and 92.5\% Rank-1.
Compared with TriPro-ReID~\cite{wang2026person},
\textbf{Paths}$^{\dagger}$ improves mAP and Rank-1 by 1.0\% and 1.4\% , respectively.
The consistent performance gains demonstrate the generalization of our method on large-scale simulated event data.

\noindent\textbf{Results on iLIDS-VID.}
As shown in Tab.~\ref{tab:robustness_results}, \textbf{Paths}$^{\dagger}$ achieves the best performance under both Blur and Occlusion settings.
Under Blur, it obtains 74.9\% mAP and 65.2\% Rank-1.
Under Occlusion, \textbf{Paths}$^{\dagger}$ achieves 88.3\% mAP and 76.4\% Rank-1, exceeding LER-ReID by 4.3\% and 1.7\%, respectively.
These results indicate that our method remains effective under blur and occlusion.
\begin{table}[!t]
  \caption{Comparison with different methods on MARS.}
  \label{tab:sota_mars}
  \centering
  \PathsTableStyle
  \begin{tabular}{cccc}
    \noalign{\hrule height 1pt}
    Method & Backbone & mAP & R-1 \\
    \hline
    GRL~\cite{liu2021watching} & ResNet50 & 82.8 & 88.7 \\
    OSNet~\cite{zhou2019omni} & ResNet50 & 81.9 & 87.7 \\
    SRS-Net~\cite{wang2020simple} & ResNet50 & 83.8 & 89.3 \\
    STMN~\cite{eom2021video} & ResNet50 & 83.4 & 89.0 \\
    CTL~\cite{liu2021} & ResNet50 & 85.3 & 89.6 \\
    PSTA~\cite{wang2021pyramid} & ResNet50 & 85.1 & 89.9 \\
    SDCL~\cite{cao2023event} & ResNet50 & 86.5 & 91.1 \\
    TriPro-ReID~\cite{wang2026person} & CLIP-B/16 & \underline{88.4} & 91.1 \\
    \hline
    \textbf{Paths}$^{\ast}$ & CLIP-B/16 & 88.1 & \underline{91.3} \\
    \rowcolor[gray]{0.92}
    \textbf{Paths}$^{\dagger}$ & DINOv3 & \textbf{89.4} & \textbf{92.5} \\
    \noalign{\hrule height 1pt}
  \end{tabular}
\end{table}
\begin{table}[!t]
  \caption{Comparison with different methods on iLIDS-VID.}
  \label{tab:robustness_results}
  \centering
  \PathsDenseTableStyle
  \footnotesize
  \setlength{\tabcolsep}{5pt}
  \begin{tabular}{cccccc}
    \noalign{\hrule height 1pt}
    Method & Backbone & Blur-mAP & Blur-R1 & Occ.-mAP & Occ.-R1 \\
    \hline
    HashReID~\cite{nikhal2023hashreid} & ResNet50 & 67.0 & 55.3 & 65.5 & 55.3 \\
    Event-ReID~\cite{cao2023event} & ResNet50 & 65.0 & 52.7 & 65.7 & 54.0 \\
    BiCnet-TKS~\cite{hou2021bicnet} & ResNet50 & 65.2 & 54.0 & 68.4 & 56.7 \\
    TCLNet~\cite{hou2020temporal} & ResNet50 & 64.1 & 51.3 & 74.6 & 64.7 \\
    SINet~\cite{bai2022salient} & ResNet50 & 55.5 & 43.3 & 69.4 & 58.0 \\
    STRF~\cite{9711297} & ResNet50 & 65.1 & 53.3 & 73.4 & 60.7 \\
    GRL~\cite{liu2021watching} & ResNet50 & 65.2 & 56.0 & 79.2 & 70.7 \\
    SRS-Net~\cite{wang2020simple} & ResNet50 & 60.8 & 48.7 & 69.0 & 58.7 \\
    STMN~\cite{eom2021video} & ResNet50 & 62.2 & 49.3 & 73.5 & 62.7 \\
    CTL~\cite{liu2021} & ResNet50 & 62.4 & 50.7 & 75.9 & 65.3 \\
    PSTA~\cite{wang2021pyramid} & ResNet50 & 62.8 & 48.0 & 77.7 & 67.3 \\
    SDCL~\cite{cao2023event} & ResNet50 & 69.2 & 58.0 & 80.7 & 71.3 \\
    LER-ReID~\cite{cao2026learning} & ResNet50 & 73.4 & 62.0 & 84.0 & 74.7 \\
    CAViT~\cite{wu2022cavit} & ViT-B/16 & 66.2 & 54.7 & 65.8 & 55.3 \\
    \hline
    \textbf{Paths}$^{\ast}$ & CLIP-B/16 & \underline{74.0} & \underline{63.1} & \underline{86.0} & \underline{75.1} \\
    \rowcolor[gray]{0.92}
    \textbf{Paths}$^{\dagger}$ & DINOv3 & \textbf{74.9} & \textbf{65.2} & \textbf{88.3} & \textbf{76.4} \\
    \noalign{\hrule height 1pt}
  \end{tabular}
\end{table}
\subsection{Ablation Study}
We conduct ablation experiments on the EvReID dataset using DINOv3 as the backbone to evaluate the contribution of Paths.
\begin{figure*}[t]
  \centering
  \includegraphics[width=1.0\textwidth]{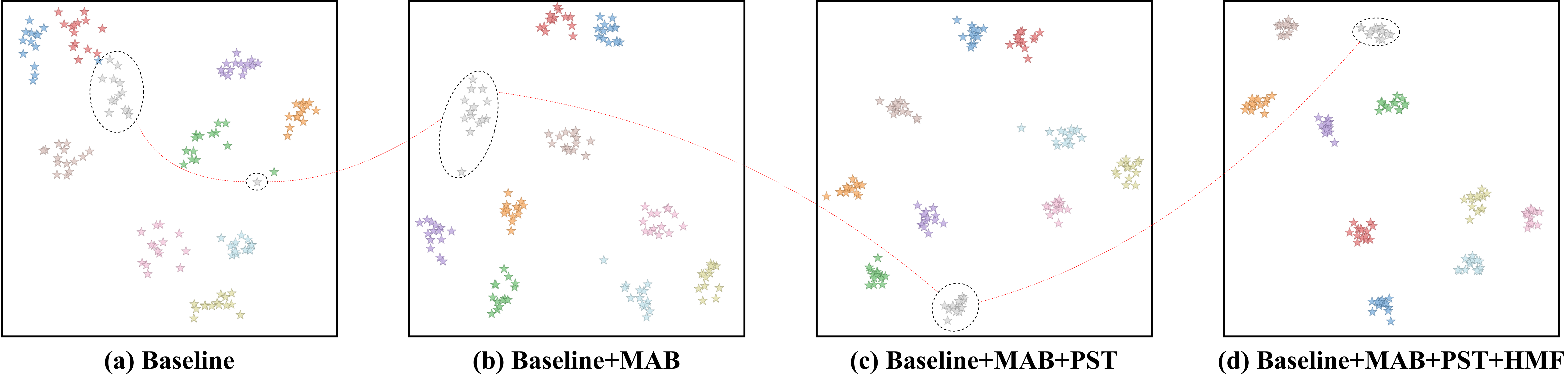}
  \vspace{-20pt}
  \caption{Visualization of the feature distributions with t-SNE. Different colors stand for different identities.}
  \label{fig:tsne_vis}
\end{figure*}

\noindent\textbf{Effects of Key Modules.}
Tab.~\ref{tab:ablation_main} reports the contribution of the main
components in our method.
MAB improves cross-batch identity supervision, while PST and HMF enhance spatio-temporal modeling and global-local interaction, respectively.
Combining all modules achieves 73.6\% mAP, 90.8\% Rank-1, 96.2\% Rank-5, and 97.8\% Rank-10, demonstrating the benefits of these components.

\noindent\textbf{Effects of PST.}
Tab.~\ref{tab:ablation_tpt} evaluates the main components of PST.
Using temporal prompts alone yields 69.9\% mAP and 89.4\% Rank-1.
Combining the TPS with BTS yields the best performance, leading to 73.6\% mAP and 90.8\% Rank-1.

\noindent\textbf{Effects of HMF.}
Tab.~\ref{tab:ablation_hca} evaluates the main components of HMF.
The global branch alone obtains 71.6\% mAP, while combining global and local branches with token matching and $\mathcal{L}_{\mathit{align}}$ achieves the best performance of 73.6\% mAP and 90.8\% Rank-1.
%
\begin{table}[!t]
  \caption{Component ablation on EvReID.}
  \label{tab:ablation_main}
  \centering
  \PathsTableStyle
  \renewcommand{\arraystretch}{0.98}
  \setlength{\tabcolsep}{3.5pt}
  \begin{tabular}{cccccccc}
    \noalign{\hrule height 1pt}
    No. & MAB & PST & HMF & mAP & R-1 & R-5 & R-10 \\
    \hline
    1 & $\times$ & $\times$ & $\times$ & 61.0 & 80.5 & 88.4 & 92.2 \\
    2 & $\checkmark$ & $\times$ & $\times$ & 63.4 & 82.3 & 90.1 & 93.5 \\
    3 & $\checkmark$ & $\checkmark$ & $\times$ & 68.8 & 86.5 & 92.4 & 95.2 \\
    4 & $\checkmark$ & $\times$ & $\checkmark$ & 69.2 & 87.8 & 94.8 & 96.6 \\
    \rowcolor[gray]{0.92}
    5 & $\checkmark$ & $\checkmark$ & $\checkmark$
      & \textbf{73.6} & \textbf{90.8}
      & \textbf{96.2} & \textbf{97.8} \\
    \noalign{\hrule height 1pt}
  \end{tabular}
\end{table}
\begin{table}[!t]
  \caption{Ablation of the PST design on EvReID.}
  \label{tab:ablation_tpt}
  \centering
  \PathsTableStyle
  \renewcommand{\arraystretch}{0.98}
  \setlength{\tabcolsep}{3.5pt}
  \begin{tabular}{cccccc}
    \noalign{\hrule height 1pt}
    No. & Prompts & TPS & BTS & mAP & R-1 \\
    \hline
    1 & $\checkmark$ & $\times$ & $\times$
      & 69.9 & 89.4 \\
    2 & $\checkmark$ & $\checkmark$ & $\times$
      & 71.5 & 89.9 \\
    3 & $\checkmark$ & $\times$ & $\checkmark$
      & 71.8 & 90.2 \\
    \rowcolor[gray]{0.92}
    4 & $\checkmark$ & $\checkmark$ & $\checkmark$
      & \textbf{73.6} & \textbf{90.8} \\
    \noalign{\hrule height 1pt}
  \end{tabular}
\end{table}
\begin{table}[!t]
  \caption{Ablation of the HMF design on EvReID.}
  \label{tab:ablation_hca}
  \centering
  \PathsTableStyle
  \renewcommand{\arraystretch}{0.98}
  \setlength{\tabcolsep}{3.5pt}
  \begin{tabular}{ccccccc}
    \noalign{\hrule height 1pt}
    No. & Global & Local & Hungarian & $\mathcal{L}_{\mathit{align}}$ & mAP & R-1 \\
    \hline
    1 & $\checkmark$ & $\times$ & $\times$ & $\times$
      & 71.6 & 89.3 \\
    2 & $\times$ & $\checkmark$ & $\checkmark$ & $\checkmark$
      & 71.1 & 88.9 \\
    3 & $\checkmark$ & $\checkmark$ & $\times$ & $\times$
      & 72.0 & 89.6 \\
    4 & $\checkmark$ & $\checkmark$ & $\checkmark$ & $\times$
      & 72.7 & 90.1 \\
    \rowcolor[gray]{0.92}
    5 & $\checkmark$ & $\checkmark$ & $\checkmark$ & $\checkmark$
      & \textbf{73.6} & \textbf{90.8} \\
    \noalign{\hrule height 1pt}
  \end{tabular}
\end{table}

\begin{figure}[t]
  \centering
  \includegraphics[width=1.0\linewidth]{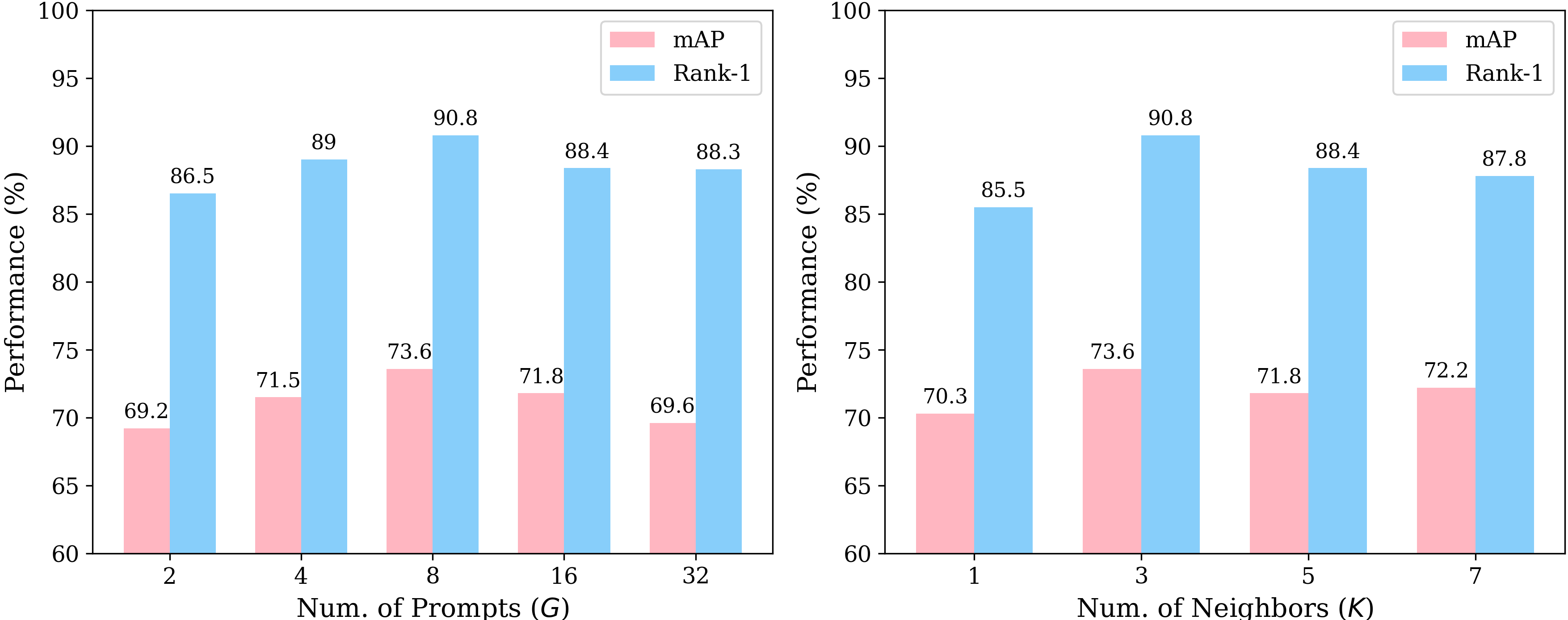}
  \vspace{-20pt}
  \caption{Effect of hyperparameters $G$ and $K$ on EvReID.}
  \label{fig:hyperparameter_analysis}
\end{figure}

\noindent\textbf{Effects of Prompt Number in PST.}
As shown in Fig.~\ref{fig:hyperparameter_analysis},
we evaluate the number of temporal prompts by $G\in\{2,4,8,16,32\}$.
The performance improves with increasing $G$, reaching its best at
$G=8$.
Thus, we set $G=8$ in all experiments.

\noindent\textbf{Effects of Neighborhood Size in HMF.}
As shown in Fig.~\ref{fig:hyperparameter_analysis}, we evaluate the number of selected neighbors $K$ in GGF from 1 to 7.
The best performance is achieved at $K=3$.
\begin{figure}[t]
  \centering
  \includegraphics[width=1.0\linewidth]{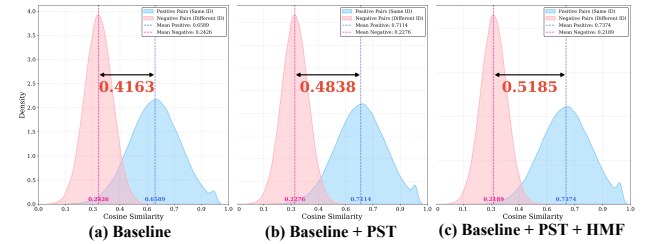}
  \vspace{-20pt}
  \caption{Visualization of the cosine similarity distribution.}
  \label{fig:cosine_similarity}
\end{figure}
\subsection{Visualization Analysis}
\noindent\textbf{Visualization of Multi-modal Feature Distributions.}
Fig.~\ref{fig:tsne_vis} presents the t-SNE of feature distributions on the test set.
Compared with the baseline, PST produces more compact identity clusters,
and HMF further strengthens feature compactness and class separation.

\noindent\textbf{Cosine Similarity Distributions.}
Fig.~\ref{fig:cosine_similarity} presents the cosine similarity
distributions of positive and negative pairs on the test set.
Compared with the baseline, PST reduces the overlap between the two distributions, and HMF further
increases the similarity margin between positive and negative pairs.
\section{Conclusion}
In this paper, we present Paths, a novel framework with spatio-temporal modeling and hierarchical multi-modal fusion for RE-VReID.
More specifically, we first propose MAB to maintain modality- specific identity prototypes for intra-modal representation learning.
Then, we introduce PST with learnable temporal prompts to jointly model spatial and temporal cues within a unified Transformer.
Finally, we use HMF to fuse RGB and event features at both global and local levels.
Extensive experiments on three RE-VReID
benchmarks demonstrate the effectiveness of our proposed method.
%
%
\bibliographystyle{ACM-Reference-Format}
\bibliography{sample-base}

@STRING{jun = "June"}

@inproceedings{li2018di,
  author    = {Li, Shuang and Bak, Slawomir and Carr, Peter and Wang, Xiaogang},
  title     = {Diversity regularized spatiotemporal attention for video-based person re-identification},
  booktitle = {Proceedings of the IEEE Conference on Computer Vision and Pattern Recognition},
  pages     = {369--378},
  year      = {2018},
}

@inproceedings{7780517,
  author    = {McLaughlin, Niall and Del Rincon, Jesus Martinez and Miller, Paul},
  title     = {Recurrent convolutional network for video-based person re-identification},
  booktitle = {Proceedings of the IEEE Conference on Computer Vision and Pattern Recognition},
  pages     = {1325--1334},
  year      = {2016},
}

@inproceedings{wang2014person,
  author    = {Wang, Taiqing and Gong, Shaogang and Zhu, Xiatian and Wang, Shengjin},
  title     = {Person re-identification by video ranking},
  booktitle = {Proceedings of the European Conference on Computer Vision},
  pages     = {688--703},
  year      = {2014},
}

@inproceedings{wang2021pyramid,
  author    = {Wang, Yingquan and Zhang, Pingping and Gao, Shang and Geng, Xia and Lu, Hu and Wang, Dong},
  title     = {Pyramid spatial-temporal aggregation for video-based person re-identification},
  booktitle = {Proceedings of the IEEE/CVF International Conference on Computer Vision},
  pages     = {12026--12035},
  year      = {2021},
}

@inproceedings{yan2020learning,
  author    = {Yan, Yichao and Qin, Jie and Chen, Jiaxin and Liu, Li and Zhu, Fan and Tai, Ying and Shao, Ling},
  title     = {Learning multi-granular hypergraphs for video-based person re-identification},
  booktitle = {Proceedings of the IEEE/CVF Conference on Computer Vision and Pattern Recognition},
  pages     = {2899--2908},
  year      = {2020},
}

@inproceedings{9711297,
  author    = {Aich, Abhishek and Zheng, Meng and Karanam, Srikrishna and Chen, Terrence and Roy-Chowdhury, Amit K and Wu, Ziyan},
  title     = {Spatio-temporal representation factorization for video-based person re-identification},
  booktitle = {Proceedings of the IEEE/CVF International Conference on Computer Vision},
  pages     = {152--162},
  year      = {2021},
}

@inproceedings{eom2021video,
  author    = {Eom, Chanho and Lee, Geon and Lee, Junghyup and Ham, Bumsub},
  title     = {Video-based person re-identification with spatial and temporal memory networks},
  booktitle = {Proceedings of the IEEE/CVF International Conference on Computer Vision},
  pages     = {12036--12045},
  year      = {2021},
}

@inproceedings{gu2020appearance,
  author    = {Gu, Xinqian and Chang, Hong and Ma, Bingpeng and Zhang, Hongkai and Chen, Xilin},
  title     = {Appearance-preserving 3d convolution for video-based person re-identification},
  booktitle = {Proceedings of the European Conference on Computer Vision},
  pages     = {228--243},
  year      = {2020},
}

@inproceedings{hou2021bicnet,
  author    = {Hou, Ruibing and Chang, Hong and Ma, Bingpeng and Huang, Rui and Shan, Shiguang},
  title     = {Bicnet-tks: Learning efficient spatial-temporal representation for video person re-identification},
  booktitle = {Proceedings of the IEEE/CVF Conference on Computer Vision and Pattern Recognition},
  pages     = {2014--2023},
  year      = {2021},
}

@inproceedings{liu2021,
  author    = {Liu, Jiawei and Zha, Zheng-Jun and Wu, Wei and Zheng, Kecheng and Sun, Qibin},
  title     = {Spatial-temporal correlation and topology learning for person re-identification in videos},
  booktitle = {Proceedings of the IEEE/CVF Conference on Computer Vision and Pattern Recognition},
  pages     = {4370--4379},
  year      = {2021},
}

@article{liu2023deeply,
  author  = {Liu, Xuehu and Yu, Chenyang and Zhang, Pingping and Lu, Huchuan},
  title   = {Deeply coupled convolution--transformer with spatial--temporal complementary learning for video-based person re-identification},
  journal = {IEEE Transactions on Neural Networks and Learning Systems},
  volume  = {35},
  number  = {10},
  pages   = {13753--13763},
  year    = {2023},
}

@inproceedings{cao2023event,
  author    = {Cao, Chengzhi and Fu, Xueyang and Liu, Hongjian and Huang, Yukun and Wang, Kunyu and Luo, Jiebo and Zha, Zheng-Jun},
  title     = {Event-guided person re-identification via sparse-dense complementary learning},
  booktitle = {Proceedings of the IEEE/CVF Conference on Computer Vision and Pattern Recognition},
  pages     = {17990--17999},
  year      = {2023},
}

@inproceedings{wang2026person,
  author    = {Wang, Xiao and Zhu, Qian and Wu, Shujuan and Jiang, Bo and Zhang, Shiliang},
  title     = {When Person Re-Identification Meets Event Camera: A Benchmark Dataset and an Attribute-Guided Re-Identification Framework},
  booktitle = {Proceedings of the AAAI Conference on Artificial Intelligence},
  pages     = {10172--10180},
  year      = {2026},
}

@article{gallego2022event,
  author  = {Gallego, Guillermo and Delbruck, Tobi and Orchard, Garrick and Bartolozzi, Chiara and Taba, Brian and Censi, Andrea and Leutenegger, Stefan and Davison, Andrew J. and Conradt, Jorg and Daniilidis, Kostas and Scaramuzza, Davide},
  title   = {Event-Based Vision: A Survey},
  journal = {IEEE Transactions on Pattern Analysis and Machine Intelligence},
  volume  = {44},
  number  = {1},
  pages   = {154--180},
  year    = {2022},
}

@inproceedings{li2025eventbasedcmtc,
  author    = {Li, Renkai and Yuan, Xin and Liu, Wei and Xu, Xin},
  title     = {Event-based Video Person Re-identification via Cross-Modality and Temporal Collaboration},
  booktitle = {Proceedings of the 2025 IEEE International Conference on Acoustics, Speech and Signal Processing},
  pages     = {1--5},
  year      = {2025},
}

@article{chen2025hierarchical,
  author  = {Chen, Wanzhang and Kong, Jun and Jiang, Min and Tao, Xuefeng},
  title   = {Hierarchical semantic compactness and proxy sparsity learning with event inversion for event person re-identification},
  journal = {Journal of Electronic Imaging},
  volume  = {34},
  number  = {2},
  pages   = {023045},
  year    = {2025},
}

@article{Tan2025SpectrumguidedFE,
  author  = {Tan, Hongchen and Zhang, Yi and Liu, Xiuping and Yin, Baocai and Ma, Nan and Li, Xin and Lu, Huchuan},
  title   = {Spectrum-Guided Feature Enhancement Network for Event Person Re-Identification},
  journal = {Pattern Recognition},
  volume  = {172},
  pages   = {112705},
  year    = {2026},
}

@article{zhou2022learningprompt,
  author  = {Zhou, Kaiyang and Yang, Jingkang and Loy, Chen Change and Liu, Ziwei},
  title   = {Learning to Prompt for Vision-Language Models},
  journal = {International Journal of Computer Vision},
  volume  = {130},
  number  = {9},
  pages   = {2337--2348},
  year    = {2022},
}

@inproceedings{jia2022visualprompt,
  author    = {Jia, Menglin and Tang, Luming and Chen, Bor-Chun and Cardie, Claire and Belongie, Serge and Hariharan, Bharath and Lim, Ser-Nam},
  title     = {Visual Prompt Tuning},
  booktitle = {Proceedings of the European Conference on Computer Vision},
  pages     = {709--727},
  year      = {2022},
}

@inproceedings{khattak2023maple,
  author    = {Khattak, Muhammad Uzair and Rasheed, Hanoona and Maaz, Muhammad and Khan, Salman and Khan, Fahad Shahbaz},
  title     = {{MaPLe}: Multi-Modal Prompt Learning},
  booktitle = {Proceedings of the IEEE/CVF Conference on Computer Vision and Pattern Recognition},
  pages     = {19113--19122},
  year      = {2023},
}

@inproceedings{li2023clip,
  author    = {Li, Siyuan and Sun, Li and Li, Qingli},
  title     = {{{CLIP}}-reid: exploiting vision-language model for image re-identification without concrete text labels},
  booktitle = {Proceedings of the AAAI Conference on Artificial Intelligence},
  pages     = {1405--1413},
  year      = {2023},
}

@inproceedings{lin2025exploringpartinformedvisuallanguagelearning,
  author    = {Lin, Yin and Chen, Yehansen and Yin, Baocai and Hu, Jinshui and Yin, Bing and Liu, Cong and Wang, Zengfu},
  title     = {Exploring part-informed visual-language learning for person re-identification},
  booktitle = {Proceedings of the 2025 IEEE International Conference on Multimedia and Expo},
  pages     = {1--6},
  year      = {2025},
}

@inproceedings{huang2023vop,
  author    = {Huang, Siteng and Gong, Biao and Pan, Yulin and Jiang, Jianwen and Lv, Yiliang and Li, Yuyuan and Wang, Donglin},
  title     = {{VoP}: Text-Video Co-Operative Prompt Tuning for Cross-Modal Retrieval},
  booktitle = {Proceedings of the IEEE/CVF Conference on Computer Vision and Pattern Recognition},
  pages     = {6565--6574},
  year      = {2023},
}

@inproceedings{liu2025stop,
  author    = {Liu, Zichen and Xu, Kunlun and Su, Bing and Zou, Xu and Peng, Yuxin and Zhou, Jiahuan},
  title     = {{STOP}: Integrated Spatial-Temporal Dynamic Prompting for Video Understanding},
  booktitle = {Proceedings of the IEEE/CVF Conference on Computer Vision and Pattern Recognition},
  pages     = {13776--13786},
  year      = {2025},
}

@inproceedings{li2026cadtrack,
  author    = {Li, Hao and Wang, Yuhao and Hu, Xiantao and Hao, Wenning and Zhang, Pingping and Wang, Dong and Lu, Huchuan},
  title     = {{CADTrack}: Learning Contextual Aggregation with Deformable Alignment for Robust {RGBT} Tracking},
  booktitle = {Proceedings of the AAAI Conference on Artificial Intelligence},
  pages     = {6109--6117},
  year      = {2026},
}

@inproceedings{li2026ragtrack,
  author    = {Li, Hao and Wang, Yuhao and Hao, Wenning and Zhang, Pingping and Wang, Dong and Lu, Huchuan},
  title     = {{RAGTrack}: Language-Aware {RGBT} Tracking with Retrieval-Augmented Generation},
  booktitle = {Proceedings of the IEEE/CVF Conference on Computer Vision and Pattern Recognition},
  pages     = {28179--28189},
  year      = {2026},
}

@article{cao2026learning,
  author  = {Cao, Chengzhi and Fu, Xueyang and Xu, Senyan and Ge, Chengjie and Wang, Kunyu and Zha, Zheng-Jun},
  title   = {Learning Robust Event-Guided Representations for Person Re-Identification},
  journal = {International Journal of Computer Vision},
  volume  = {134},
  number  = {2},
  pages   = {82},
  year    = {2026},
}

@inproceedings{yu2026xreid,
  author    = {Yu, Chenyang and Liu, Xuehu and Zhang, Pingping and Lu, Huchuan},
  title     = {{X-ReID}: Multi-granularity Information Interaction for Video-Based Visible-Infrared Person Re-Identification},
  booktitle = {Proceedings of the AAAI Conference on Artificial Intelligence},
  pages     = {12117--12125},
  year      = {2026},
}

@article{Hungarian,
  author  = {Kuhn, H. W.},
  title   = {The Hungarian method for the assignment problem},
  journal = {Naval Research Logistics Quarterly},
  volume  = {2},
  number  = {1-2},
  pages   = {83--97},
  year    = {1955},
}

@inproceedings{zhou2019omni,
  author    = {Zhou, Kaiyang and Yang, Yongxin and Cavallaro, Andrea and Xiang, Tao},
  title     = {Omni-scale feature learning for person re-identification},
  booktitle = {Proceedings of the IEEE/CVF International Conference on Computer Vision},
  pages     = {3702--3712},
  year      = {2019},
}

@inproceedings{hou2020temporal,
  author    = {Hou, Ruibing and Chang, Hong and Ma, Bingpeng and Shan, Shiguang and Chen, Xilin},
  title     = {Temporal complementary learning for video person re-identification},
  booktitle = {Proceedings of the European Conference on Computer Vision},
  pages     = {388--405},
  year      = {2020},
}

@inproceedings{liu2021watching,
  author    = {Liu, Xuehu and Zhang, Pingping and Yu, Chenyang and Lu, Huchuan and Yang, Xiaoyun},
  title     = {Watching you: Global-guided reciprocal learning for video-based person re-identification},
  booktitle = {Proceedings of the IEEE/CVF Conference on Computer Vision and Pattern Recognition},
  pages     = {13334--13343},
  year      = {2021},
}

@inproceedings{bai2022salient,
  author    = {Bai, Shutao and Ma, Bingpeng and Chang, Hong and Huang, Rui and Chen, Xilin},
  title     = {Salient-to-broad transition for video person re-identification},
  booktitle = {Proceedings of the IEEE/CVF Conference on Computer Vision and Pattern Recognition},
  pages     = {7339--7348},
  year      = {2022},
}

@inproceedings{yu2024tf,
  author    = {Yu, Chenyang and Liu, Xuehu and Wang, Yingquan and Zhang, Pingping and Lu, Huchuan},
  title     = {Tf-clip: Learning text-free clip for video-based person re-identification},
  booktitle = {Proceedings of the AAAI Conference on Artificial Intelligence},
  pages     = {6764--6772},
  year      = {2024},
}

@inproceedings{yu2025climb,
  author    = {Yu, Chenyang and Liu, Xuehu and Zhu, Jiawen and Wang, Yuhao and Zhang, Pingping and Lu, Huchuan},
  title     = {Climb-reid: A hybrid clip-mamba framework for person re-identification},
  booktitle = {Proceedings of the AAAI Conference on Artificial Intelligence},
  pages     = {9589--9597},
  year      = {2025},
}

@inproceedings{zheng2016mars,
  author    = {Zheng, Liang and Bie, Zhi and Sun, Yifan and Wang, Jingdong and Su, Chi and Wang, Shengjin and Tian, Qi},
  title     = {{MARS}: A video benchmark for large-scale person re-identification},
  booktitle = {Proceedings of the European Conference on Computer Vision},
  pages     = {868--884},
  year      = {2016},
}

@inproceedings{dosovitskiy2020image,
  author    = {Dosovitskiy, Alexey and Beyer, Lucas and Kolesnikov, Alexander and Weissenborn, Dirk and Zhai, Xiaohua and Unterthiner, Thomas and Dehghani, Mostafa and Minderer, Matthias and Heigold, Georg and Gelly, Sylvain and Uszkoreit, Jakob and Houlsby, Neil},
  title     = {An Image Is Worth 16x16 Words: {Transformers} for Image Recognition at Scale},
  booktitle = {International Conference on Learning Representations},
  year      = {2021},
}

@article{simeoni2025dinov3,
  author  = {Sim{\'e}oni, Oriane and Vo, Huy V. and Seitzer, Maximilian and Baldassarre, Federico and Oquab, Maxime and Jose, Cijo and Khalidov, Vasil and Szafraniec, Marc and Yi, Seungeun and Ramamonjisoa, Micha{\"e}l and others},
  title   = {{DINOv3}},
  journal = {arXiv preprint arXiv:2508.10104},
  year    = {2025},
}

@inproceedings{zhong2017randomerasingdataaugmentation,
  author    = {Zhong, Zhun and Zheng, Liang and Kang, Guoliang and Li, Shaozi and Yang, Yi},
  title     = {Random erasing data augmentation},
  booktitle = {Proceedings of the AAAI Conference on Artificial Intelligence},
  pages     = {13001--13008},
  year      = {2020},
}

@article{Kingma2014AdamAM,
  author  = {Kingma, Diederik P. and Ba, Jimmy},
  title   = {{Adam}: A Method for Stochastic Optimization},
  journal = {arXiv preprint arXiv:1412.6980},
  year    = {2014},
}

@article{wang2020simple,
  author  = {Wang, Haoran and Jiao, Licheng and Yang, Shuyuan and Li, Lingling and Wang, Zexin},
  title   = {Simple and Effective: Spatial Rescaling for Person Reidentification},
  journal = {IEEE Transactions on Neural Networks and Learning Systems},
  volume  = {33},
  number  = {1},
  pages   = {145--156},
  year    = {2022},
}

@inproceedings{nikhal2023hashreid,
  author    = {Nikhal, Kshitij and Ma, Yujunrong and Bhattacharyya, Shuvra S and Riggan, Benjamin S},
  title     = {{HashReID}: Dynamic network with binary codes for efficient person re-identification},
  booktitle = {Proceedings of the IEEE/CVF Winter Conference on Applications of Computer Vision},
  pages     = {6046--6055},
  year      = {2024},
}

@inproceedings{wu2022cavit,
  author    = {Wu, Jinlin and He, Lingxiao and Liu, Wu and Yang, Yang and Lei, Zhen and Mei, Tao and Li, Stan Z},
  title     = {{Cavit}: Contextual alignment vision transformer for video object re-identification},
  booktitle = {Proceedings of the European Conference on Computer Vision},
  pages     = {549--566},
  year      = {2022},
}

@inproceedings{wang2024top,
  author    = {Wang, Yuhao and Liu, Xuehu and Zhang, Pingping and Lu, Hu and Tu, Zhengzheng and Lu, Huchuan},
  title     = {TOP-{ReID}: Multi-Spectral Object Re-identification with Token Permutation},
  booktitle = {Proceedings of the AAAI Conference on Artificial Intelligence},
  pages     = {5758--5766},
  year      = {2024},
}

@inproceedings{zhang2024magictokens,
  author    = {Zhang, Pingping and Wang, Yuhao and Liu, Yang and Tu, Zhengzheng and Lu, Huchuan},
  title     = {Magic Tokens: Select Diverse Tokens for Multi-modal Object Re-Identification},
  booktitle = {Proceedings of the IEEE/CVF Conference on Computer Vision and Pattern Recognition},
  pages     = {17117--17126},
  year      = {2024},
}

@inproceedings{wang2025demo,
  author    = {Wang, Yuhao and Liu, Yang and Zheng, Aihua and Zhang, Pingping},
  title     = {{DeMo}: Decoupled Feature-Based Mixture of Experts for Multi-Modal Object Re-Identification},
  booktitle = {Proceedings of the AAAI Conference on Artificial Intelligence},
  pages     = {8141--8149},
  year      = {2025},
}

@inproceedings{wang2025mambapro,
  author    = {Wang, Yuhao and Liu, Xuehu and Yan, Tianyu and Liu, Yang and Zheng, Aihua and Zhang, Pingping and Lu, Huchuan},
  title     = {{MambaPro}: Multi-Modal Object Re-identification with Mamba Aggregation and Synergistic Prompt},
  booktitle = {Proceedings of the AAAI Conference on Artificial Intelligence},
  pages     = {8150--8158},
  year      = {2025},
}

@inproceedings{wang2025idea,
  author    = {Wang, Yuhao and Lv, Yongfeng and Zhang, Pingping and Lu, Huchuan},
  title     = {{IDEA}: Inverted Text with Cooperative Deformable Aggregation for Multi-modal Object Re-Identification},
  booktitle = {Proceedings of the IEEE/CVF Conference on Computer Vision and Pattern Recognition},
  pages     = {29701--29710},
  year      = {2025},
}

@inproceedings{liu2026signal,
  author    = {Liu, Yangyang and Wang, Yuhao and Zhang, Pingping},
  title     = {Signal: Selective Interaction and Global-local Alignment for Multi-Modal Object Re-Identification},
  booktitle = {Proceedings of the AAAI Conference on Artificial Intelligence},
  pages     = {7359--7367},
  year      = {2026},
}

@inproceedings{yang2026hihr,
  author    = {Yang, Qiwei and Zhang, Pingping},
  title     = {{HiHR}: Hierarchical Hyperbolic Representation for Aerial-Ground Person Re-Identification},
  booktitle = {European Conference on Computer Vision},
  year      = {2026}
}

@article{wang2026sdreid,
  author  = {Wang, Yuhao and Hu, Xiang and Wang, Lixin and Zhang, Pingping and Lu, Huchuan},
  title   = {{SD-ReID}: View-Aware Stable Diffusion for Aerial-Ground Person Re-Identification},
  journal = {IEEE Transactions on Image Processing},
  volume  = {35},
  pages   = {5686--5697},
  year    = {2026}
}

@article{hu2025latex,
  author  = {Hu, Xiang and Wang, Yuhao and Zhang, Pingping and Lu, Huchuan},
  title   = {{LATex}: Leveraging Attribute-based Text Knowledge for Aerial-Ground Person Re-Identification},
  journal = {arXiv preprint arXiv:2503.23722},
  year    = {2025}
}

@article{dai2019video,
  title={Video Person Re-Identification by Temporal Residual Learning},
  author={Dai, Ju and Zhang, Pingping and Wang, Dong and Lu, Huchuan and Wang, Hongyu},
  journal={IEEE Transactions on Image Processing},
  volume={28},
  number={3},
  pages={1366--1377},
  year={2019}
}

@inproceedings{liu2023long,
  title={Video-Based Person Re-Identification with Long Short-Term Representation Learning},
  author={Liu, Xuehu and Zhang, Pingping and Lu, Huchuan},
  booktitle={Image and Graphics},
  series={Lecture Notes in Computer Science},
  volume={14355},
  pages={55--67},
  publisher={Springer},
  year={2023}
}

@article{liu2024video,
  title={A Video Is Worth Three Views: Trigeminal Transformers for Video-Based Person Re-Identification},
  author={Liu, Xuehu and Zhang, Pingping and Yu, Chenyang and Qian, Xuesheng and Yang, Xiaoyun and Lu, Huchuan},
  journal={IEEE Transactions on Intelligent Transportation Systems},
  volume={25},
  number={9},
  pages={12818--12828},
  year={2024}
}

@inproceedings{yang2026sasvpreid,
  title={{SAS-VPReID}: A Scale-Adaptive Framework with Shape Priors for Video-Based Person Re-Identification at Extreme Far Distances},
  author={Yang, Qiwei and Zhang, Pingping and Wang, Yuhao and Gong, Zijing},
  booktitle={Proceedings of the IEEE/CVF Winter Conference on Applications of Computer Vision Workshops},
  pages={1599--1608},
  year={2026}
}

@inproceedings{nguyen2025agvpreid,
  title={{AG-VPReID} 2025: Aerial-Ground Video-Based Person Re-Identification Challenge Results},
  author={Nguyen, Kien and Fookes, Clinton and Sridharan, Sridha and Nguyen, Huy and Liu, Feng and Liu, Xiaoming and Ross, Arun and Michalski, Dana and Endrei, Tamas and DeAndres-Tame, Ivan and Tolosana, Ruben and Vera-Rodriguez, Ruben and Morales, Aythami and Fierrez, Julian and Ortega-Garcia, Javier and Gong, Zijing and Wang, Yuhao and Liu, Xuehu and Zhang, Pingping and Rashidunnabi, Md and Proenca, Hugo and Hambarde, Kailash A. and Rezaei, Saeid},
  booktitle={2025 IEEE International Joint Conference on Biometrics},
  pages={1--10},
  year={2025},
  organization={IEEE}
}

@inproceedings{hambarde2026vreidxfd,
  title={{VReID-XFD}: Video-Based Person Re-Identification at Extreme Far Distance Challenge Results},
  author={Hambarde, Kailash A. and Proenca, Hugo and Rashidunnabi, Md and Samale, Pranita and Yang, Qiwei and Zhang, Pingping and Gong, Zijing and Wang, Yuhao and Zhang, Xi and Qu, Ruoshui and He, Qiaoyun and Zhang, Yuhang and Nguyen, Thi Ngoc Ha and Mai, Tien-Dung and Kang, Cheng-Jun and Lin, Yu-Fan and Jiang, Jin-Hui and Hsu, Chih-Chung and Endrei, Tamas and Cserey, Gyorgy and Rajbhandari, Ashwat},
  booktitle={Proceedings of the IEEE/CVF Winter Conference on Applications of Computer Vision Workshops},
  year={2026}
}

@article{wang2022interact,
  author  = {Wang, Zi and Li, Chenglong and Zheng, Aihua and He, Ran and Tang, Jin},
  title   = {Interact, Embed, and EnlargE: Boosting Modality-Specific Representations for Multi-Modal Person Re-identification},
  journal = {Proceedings of the AAAI Conference on Artificial Intelligence},
  volume  = {36},
  number  = {3},
  pages   = {2633--2641},
  year    = {2022}
}

@inproceedings{li2024all,
  author    = {Li, He and Ye, Mang and Zhang, Ming and Du, Bo},
  title     = {All in One Framework for Multimodal Re-identification in the Wild},
  booktitle = {Proceedings of the IEEE/CVF Conference on Computer Vision and Pattern Recognition},
  pages     = {17459--17469},
  year      = {2024}
}

@inproceedings{lin2025dmpt,
  author    = {Lin, Minghui and Wang, Shu and Wang, Xiang and Tang, Jianhua and Fu, Longbin and Zuo, Zhengrong and Sang, Nong},
  title     = {{DMPT}: Decoupled Modality-Aware Prompt Tuning for Multi-Modal Object Re-Identification},
  booktitle = {Proceedings of the IEEE/CVF Winter Conference on Applications of Computer Vision},
  pages     = {2103--2112},
  year      = {2025}
}

@inproceedings{ha2025multimodal,
  author    = {Ha, Ruiyang and Jiang, Songyi and Li, Bin and Pan, Bikang and Zhu, Yihang and Zhang, Junjie and Zhu, Xiatian and Gong, Shaogang and Wang, Jingya},
  title     = {Multi-modal Multi-platform Person Re-Identification: Benchmark and Method},
  booktitle = {Proceedings of the IEEE/CVF International Conference on Computer Vision},
  pages     = {10251--10261},
  year      = {2025}
}

@article{wang2026unique,
  title={What Makes You Unique? Attribute Prompt Composition for Object Re-Identification},
  author={Wang, Yingquan and Zhang, Pingping and Sun, Chong and Wang, Dong and Lu, Huchuan},
  journal={IEEE Transactions on Circuits and Systems for Video Technology},
  volume={36},
  number={3},
  pages={3173--3184},
  year={2026},
}

@inproceedings{szegedy2016rethinking,
  title={Rethinking the Inception Architecture for Computer Vision},
  author={Szegedy, Christian and Vanhoucke, Vincent and Ioffe, Sergey and Shlens, Jonathon and Wojna, Zbigniew},
  booktitle={Proceedings of the IEEE Conference on Computer Vision and Pattern Recognition},
  pages={2818--2826},
  year={2016}
}

@article{hermans2017defense,
  title={In Defense of the Triplet Loss for Person Re-Identification},
  author={Hermans, Alexander and Beyer, Lucas and Leibe, Bastian},
  journal={arXiv preprint arXiv:1703.07737},
  year={2017}
}
\end{document}